\documentclass[letterpaper, 10 pt, conference]{ieeeconf}  

\IEEEoverridecommandlockouts                              

\usepackage{amsmath} 
\usepackage{amssymb}  
\usepackage{mathrsfs}
\usepackage{xcolor}
\usepackage{multirow}
\usepackage{algpseudocode}
\usepackage{algorithmicx}
\usepackage[linesnumbered, ruled, vlined]{algorithm2e}
\usepackage[font=footnotesize]{caption}
\usepackage{subcaption}
\usepackage{cite}
\usepackage{graphicx}
\usepackage{dblfloatfix}
\usepackage{balance}
\usepackage{makecell}
\usepackage{booktabs}
\usepackage{multirow}
\usepackage{subcaption} 

\newtheorem{remark}{Remark}

\title{\LARGE \bf
Opinion-Driven Decision-Making for \\ Multi-Robot Navigation through Narrow Corridors
}

\author{Norah K. Alghamdi and Shinkyu Park
\thanks{This work was supported by funding from King Abdullah University of Science and Technology (KAUST)}
\thanks{The authors are with Electrical and Computer Engineering, King Abdullah University of Science and Technology (KAUST), Thuwal 23955, Saudi Arabia. {\tt \{norah.ghamdi, shinkyu.park\}@kaust.edu.sa}}
}

\begin{document}

\maketitle
\thispagestyle{empty}
\pagestyle{empty}

\begin{abstract}
We propose an opinion-driven navigation framework for multi-robot traversal through a narrow corridor. Our approach leverages a multi-agent decision-making model known as the \textit{Nonlinear Opinion Dynamics (NOD)} to address the narrow corridor passage problem, formulated as a multi-robot navigation game. By integrating the NOD model with a multi-robot path planning algorithm, we demonstrate that the framework effectively reduces the likelihood of deadlocks during corridor traversal. To ensure scalability with an increasing number of robots, we introduce a \textit{game reduction} technique that enables efficient coordination in larger groups. Extensive simulation studies are conducted to validate the effectiveness of the proposed approach.
\end{abstract}

\section{Introduction}

\begin{figure}[h]
    \centering
    \includegraphics[scale=0.34]{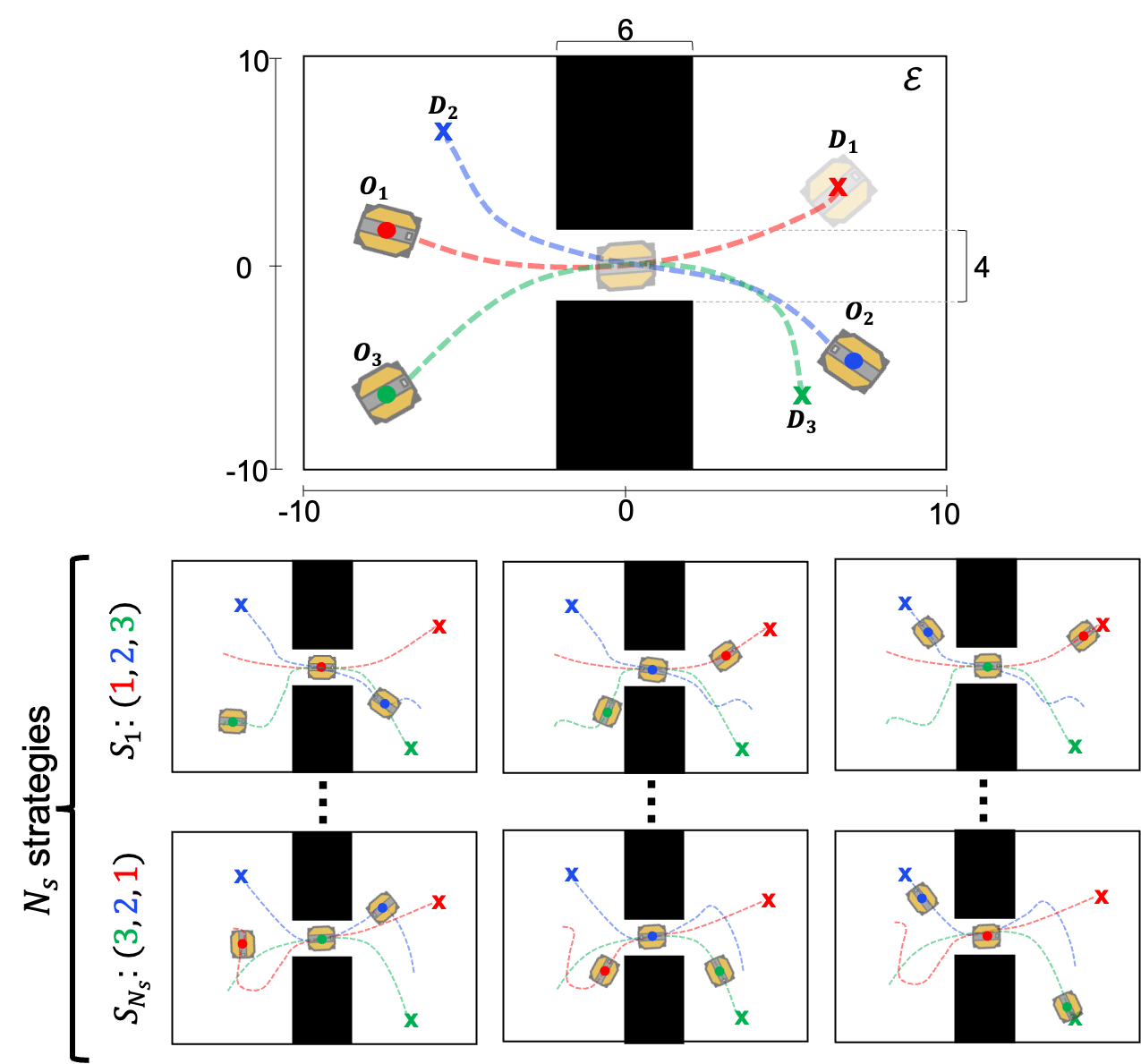}
    \caption{An Illustration of the Multi-Robot Narrow Corridor Passage: 
    Multiple strategies are defined for individual robots to adopt. Each strategy specifies the order in which the robots pass through the corridor. The proposed opinion-driven framework facilitates coordination by enabling all robots to converge on the same strategy through interactions.
    }
    \label{fig:env_pic}
    \vspace{-1.5em}
\end{figure}

In recent years, robots have become increasingly integrated into human environments, including residential areas, healthcare facilities, and public spaces. As robots interact frequently with both humans and other robots, the importance of social navigation has grown significantly. Social navigation focuses on optimizing a robot's behavior to enhance human comfort and improve the acceptability of robots in shared spaces. For instance, when multiple robots navigate through a narrow corridor, as depicted in Fig.~\ref{fig:env_pic}, they must dynamically adapt their movements in response to the actions of others, ensuring smooth and cooperative interactions within such constrained environments.

We investigate a particular social navigation scenario where multiple agents must traverse a narrow corridor without relying on centralized coordination or explicit communication. While this work primarily focuses on multi-robot navigation, the absence of explicit communication is particularly critical in scenarios where human pedestrians are also present as agents. In such situations, uncoordinated or self-centered navigation behavior can obstruct the corridor, leading to potential deadlocks. To address this difficulty, we introduce a decision-making framework that enables robots to infer and align their navigation strategies with others, based on their own perceptions of the intentions underlying the observed movements of other agents.

Our approach formulates the problem as a multi-robot navigation game -- representing strategic interactions among multiple robots navigating through a narrow corridor -- while leveraging the \textit{Nonlinear Opinion Dynamics (NOD)} model. This model, introduced in \cite{nonlinear_opinion_dynamics, bizyaeva2024multitopicbeliefformationbifurcations} and applied in multi-agent games \cite{tuning_reciprocity}, allows robots to form and dynamically update opinions about their available navigation strategies by observing and estimating the opinions of others over time. These evolving opinions drive their decision-making, enabling them to select strategies that ensure effective coordination with other robots and achieve deadlock-free navigation.

Significant research in social navigation has focused on developing algorithms that enable robots to navigate safely through challenging and dynamic environments while minimizing disruption to human activities (see survey articles  \cite{moller2021survey, gao2022evaluation, mavrogiannis2022social, mavrogiannis2021core}). In shared environments with other mobile robots or pedestrians, a key challenge in robot navigation is resolving conflicts that could result in collision or deadlock (see survey articles \cite{mirsky2021prevention, mirsky2021conflict}). Humans typically follow well-established social conventions related to personal space \cite{burgoon1976toward}. Consequently, effective robot navigation in crowded environments requires the integration of social conventions and the prediction of pedestrian and other robot behaviors into the robot's navigation algorithm. For instance, \cite{daza2021approach} introduces a navigation strategy that integrates classical navigation algorithms with proxemic theory -- the study of how people perceive and use space for communication.

Additionally, a substantial body of work focuses on mimicking human navigation behaviors by leveraging human demonstration data and data-driven methods. Approaches such as those in  \cite{kollmitz2020learning, baghi2021sample, okal2016learning} generate trajectories that replicate observed human motions. In \cite{baghi2022sesno}, the authors propose a deep inverse reinforcement learning-based framework that trains robotic agents to learn social navigation by observing human behaviors and interactions in shared spaces. Furthermore, the algorithm presented in \cite{bera2019emotionally} incorporates pedestrians' emotional states -- deduced from facial expressions and trajectories -- to enable robots to navigate in a socially appropriate manner, ensuring smoother interactions in shared environments.

Narrow corridor passage is particularly prone to conflicts among multiple robots, presenting unique coordination challenges. For example, \cite{kaiser2019make} incorporated kinesic courtesy cues into robot motions, enhancing legibility and reducing disruptions in human-robot interactions. This approach not only reduces conflicts in doorway scenarios but also improves the efficiency of both humans and robots in reaching their respective goals. Similarly, \cite{thomas2018after} introduces ``assertive'' robot behaviors to resolve navigational deadlocks in doorways, facilitating smoother interactions in both human-robot and robot-robot settings. Although these studies are limited to two-agent scenarios (e.g., human-human, robot-human, or robot-robot) and rely on simplified, predefined navigation behaviors that may not scale directly to multi-agent corridor environments, they highlight the importance of signaling and inferring intentions. The latter, in particular, is a central component of our approach for resolving conflicts in narrow corridor navigation.

In our framework, robots observe the positions and velocities of other robots to infer their intentions -- specifically, whether a robot is attempting to enter the corridor or yielding to others. Based on this information and using the NOD model, each robot selects navigation strategies that aim to prevent deadlocks. Relevant to our work, \cite{Galati2022} proposed a game-theoretic trajectory planning method designed to improve the social acceptability of robots when interacting with humans. However, unlike our framework, their approach defines strategies based on a constant linear velocity and time-varying heading, which makes it unsuitable for the corridor navigation scenarios considered in our study. The NOD model has also been previously applied in \cite{10341745} to develop a robot navigation algorithm. In contrast to our approach, their method controls only the robot's heading direction in open environments without corridors.

The paper is organized as follows: In \S\ref{approach}, we describe our opinion-driven robot navigation framework which is based on the NOD model. In \S\ref{sec:game_reduction}, we discuss how our framework scales, particularly addressing the computational demands that increases with the number of robots, and introduce a technique to mitigate this complexity. In \S\ref{simulation_results}, we report results from extensive simulations with multiple mobile robots in narrow corridor navigation, demonstrating the effectiveness of our framework. Finally, we conclude the paper with a summary and future directions in \S\ref{conclusion}.


\section{Opinion-Driven Robot Navigation} \label{approach}

\begin{figure}
    \centering
    \includegraphics[scale=0.34]{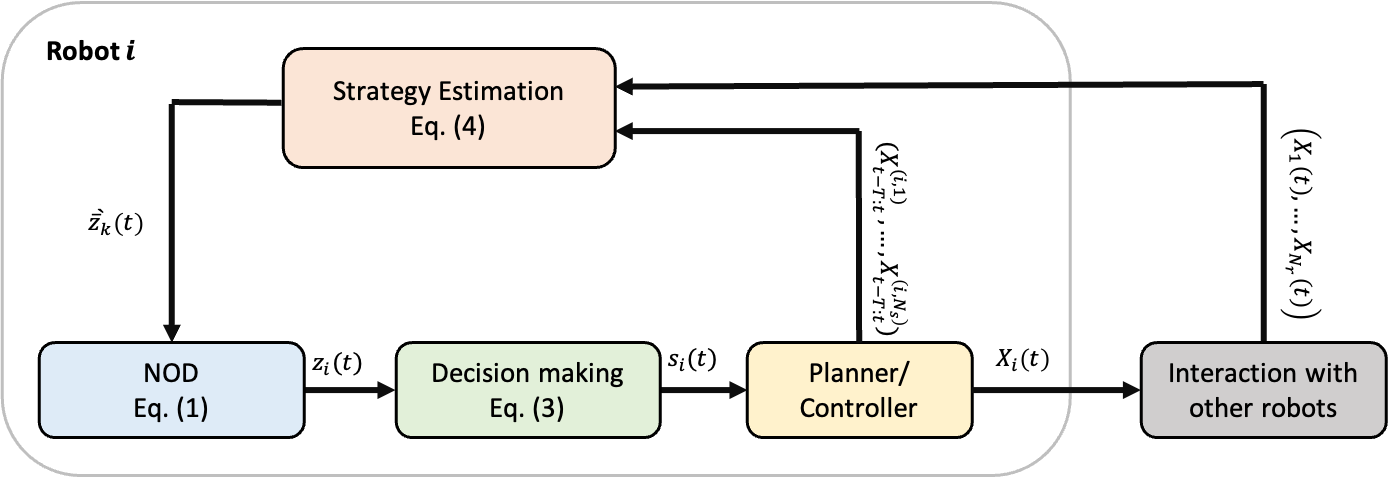}
    \caption{Framework Overview: The NOD model is designed to promote coordination during corridor navigation by aggregating incentives based on the estimated strategies of other robots, inferred from their observed motions and predicted paths provided by a multi-robot path planner. Using the updated opinion state generated by the NOD model, each robot decides on an appropriate navigation strategy. This decision is then executed through a multi-robot path planner and a motion controller, which guide the robot toward its destination.}
    \label{fig:System_overview}
    \vspace{-1.5em}
\end{figure}

We start by formally defining the narrow corridor navigation problem and then introduce the proposed framework designed to tackle its challenges. The framework mainly incorporates the Nonlinear Opinion Dynamics (NOD) model and a multi-robot path planner along with associated navigation strategies that enable robots to effectively navigate through the narrow corridor. An overview of the proposed framework is illustrated in Fig.~\ref{fig:System_overview}.

\subsection{Multi-Robot Narrow Corridor Navigation}
We address an $N_r$-robot navigation problem in a shared environment $\mathscr{E} \subset \mathbb R^2$, which includes a narrow corridor, permitting only one robot to pass at a time, as illustrated in Fig.~\ref{fig:env_pic}. Each robot $i \in \{1, \cdots, N_r\}$ aims to travel from its origin $O_{i} \in \mathscr{E}$ to its destination $D_{i} \in \mathscr{E}$, navigating through the collision-free area $\mathscr E$ while avoiding deadlocks near the narrow corridor. With $N_r$ robots, there are $N_s = {N_r}!$ unique passing orders for navigating through the corridor. Deadlocks occur when two robots attempt to pass through the narrow corridor simultaneously. The central challenge addressed in this work is determining how these $N_r$ robots can reach a consensus on the order in which they pass through the corridor. The key technical difficulty lies in the fact that the robots cannot explicitly communicate with one another; instead, they must infer each other's intentions, whether to proceed first or yield, from their observed motions.

To address the main problem, we propose an opinion-driven robot navigation framework. This framework leverages the NOD model, enabling each robot to form opinions (preferences) about the available navigation strategies. These opinions are shaped by observing the motion of other robots. Based on the formed opinions, each robot selects a navigation strategy, facilitating coordinated and efficient navigation.

\subsection{Opinion-Driven Navigation Framework}

\subsubsection{Navigation Strategy and Multi-Robot Path Planning} \label{strategy_definitioN_rnd_path_planning}

The state $X_i(t)$ of each robot~$i$ at time $t$ is defined as a combination of its position $p_i (t) \in \mathbb R^2$ and velocity $v_i (t) \in \mathbb R^2$: $X_i(t) = (p_i(t), v_i(t))$. A \textit{navigation strategy} is formally defined as one of the $N_s$ options, each specifying the order in which robots pass through the corridor to reach their respective destinations. Given a selected strategy~$j \in \{1, \cdots, N_s\}$, each robot~$i$ employs a multi-robot path planner to compute the joint path $(X_1^{(i,j)}(t), \cdots, X_{N_r}^{(i,j)}(t)), \, t \in [0, T_f]$ for all $N_r$ robots. The resulting path satisfies constraints to enforce adherence to the specified strategy and to ensure collision avoidance. Once the multi-robot path is generated, each robot~$i$ follows it to navigate toward its destination $D_i$ using a motion controller. During execution, the robot also uses the computed paths to infer the strategy selections of the other robots.\footnote{To compute accurate predictions of other robots' strategy selection, each robot is assume to know the destinations of others.}

\begin{figure}
    \centering
    \begin{subfigure}[b]{0.3\columnwidth}
        \includegraphics[width=\textwidth]{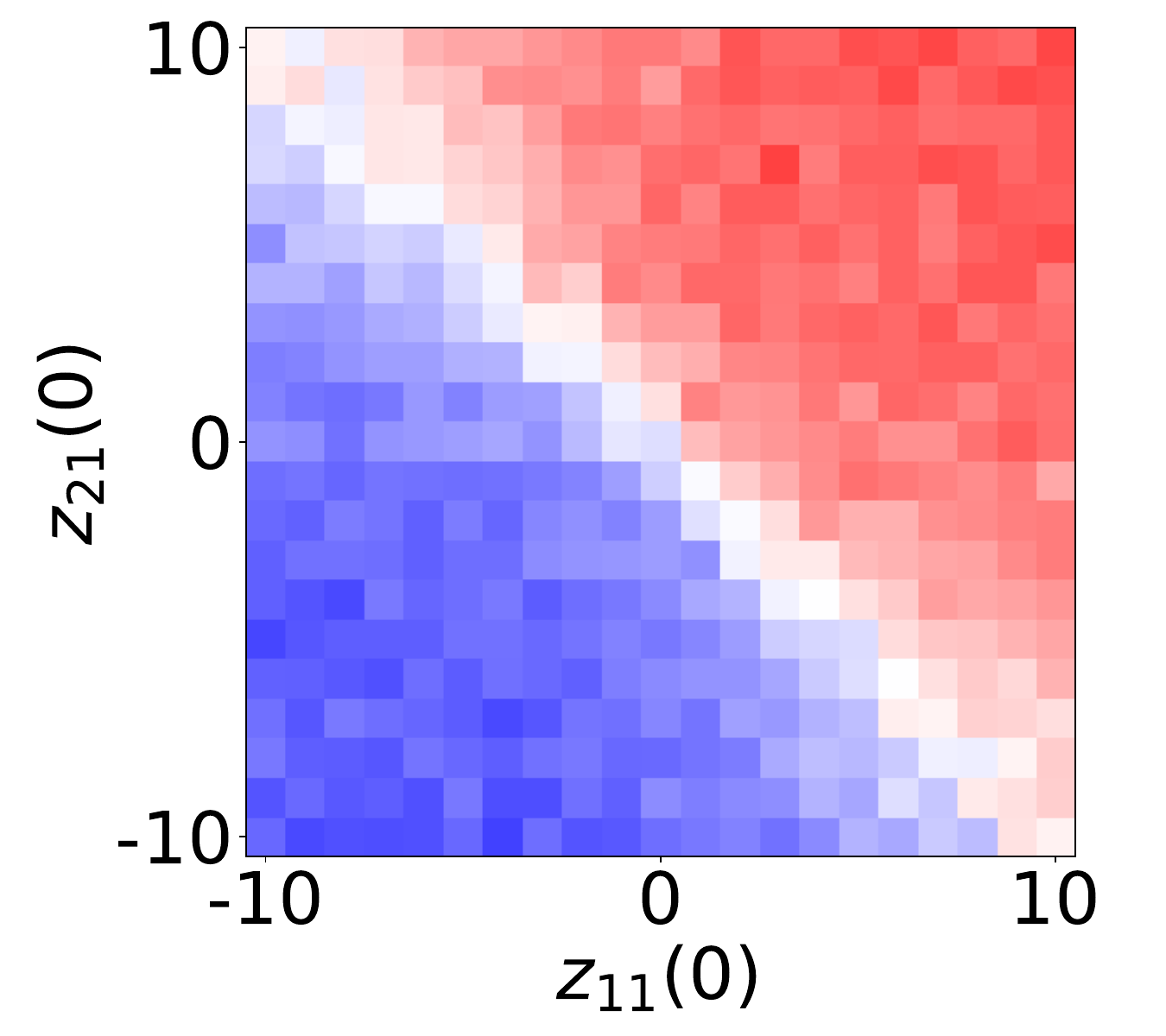}
        \caption{$ u_i = 1$}
    \end{subfigure}
    \begin{subfigure}[b]{0.3\columnwidth}
        \includegraphics[width=\textwidth]{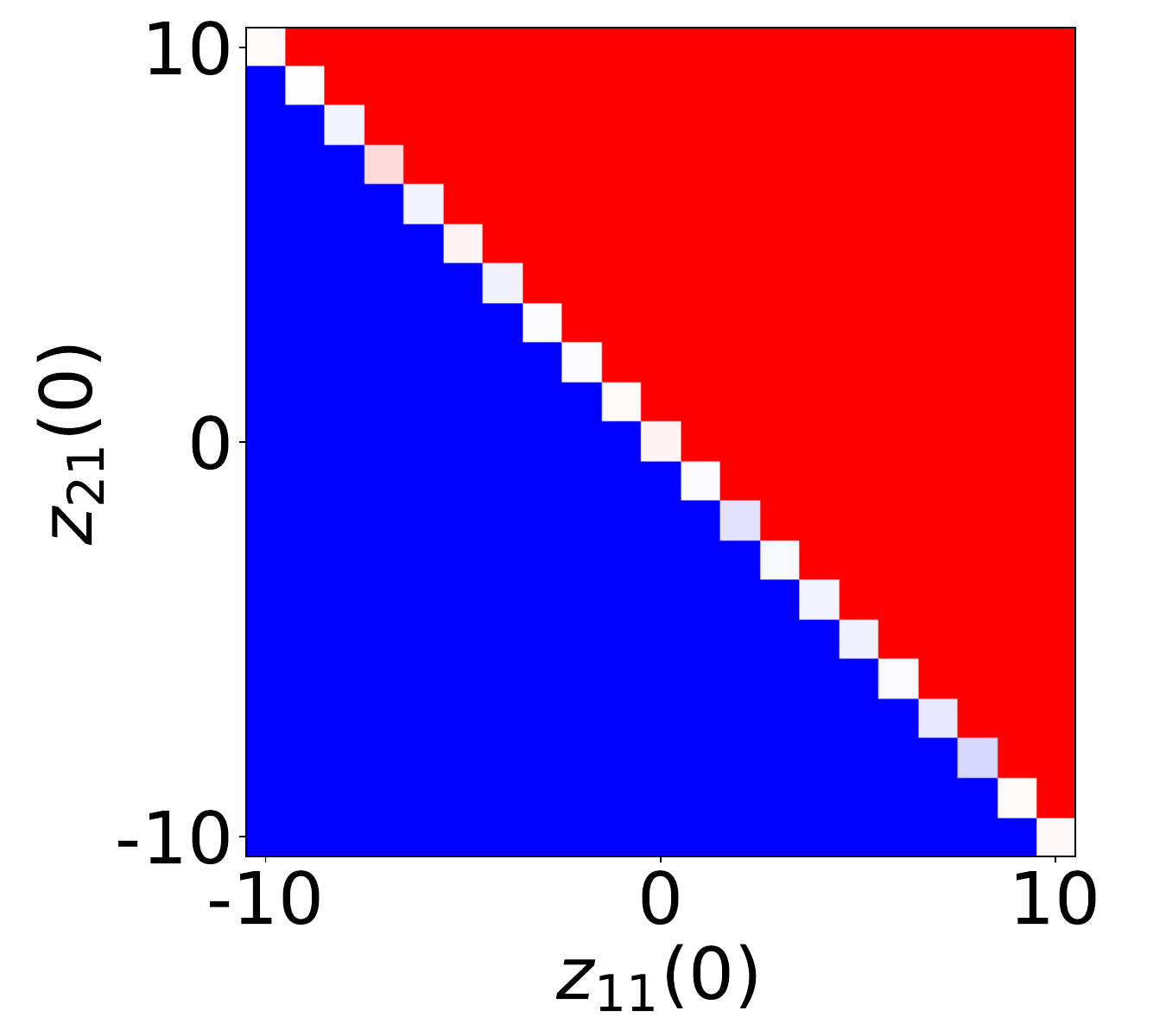}
        \caption{$u_i = 10$}
        \label{fig:2D_heatmap_b}
    \end{subfigure}
    \begin{subfigure}[b]{0.34\columnwidth}
        \includegraphics[width=\textwidth]{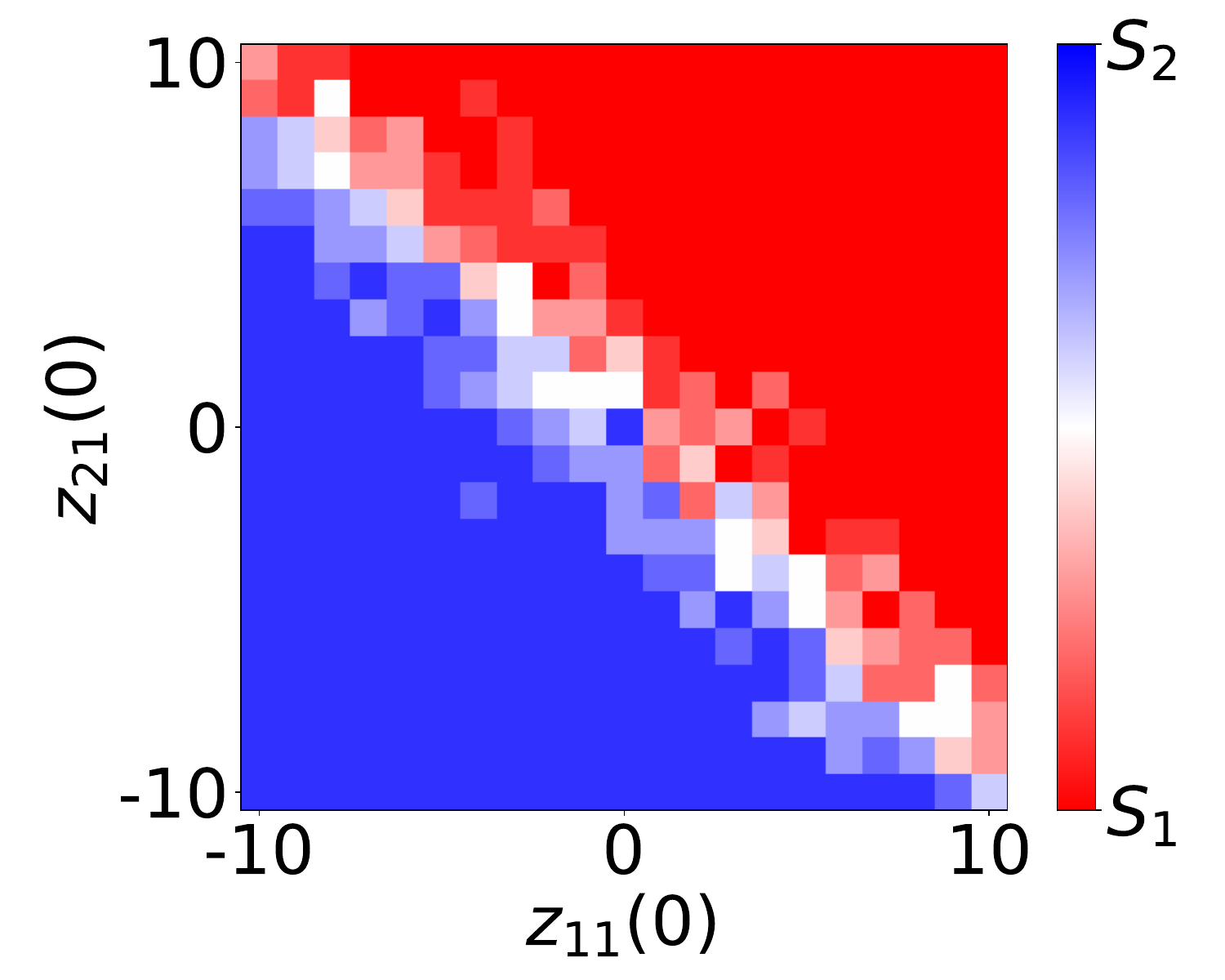}
        \caption{$u_i = 10$}
        \label{fig:2D_heatmap_c}
    \end{subfigure}
    \caption{Heatmap plots illustrate the strategy selections in a two-robot scenario with two possible strategies -- $S_1:(1,2)$ and $S_2:(2,1)$. The initial condition $z_{i1} (0)$ varies within the range $[-10, 10]$, while $z_{i2}(0) = 0$ is fixed for both robots ($i \in \{1,2\}$). Plots (a) and (b) show the strategy selections based on the steady-state limit $\lim_{t \to \infty} x_{11}(t)$, which is identical to $\lim_{t \to \infty} x_{21}(t)$, for $u_i=1$ and $u_i = 10$, respectively. These results are averaged over multiple simulations and computed directly using \eqref{eq:opinion_dynamics_continous} and \eqref{eq:strategy_selection_softmax}. Plot (c) depicts the strategy selections of both robots averaged over multiple simulations using the opinion-driven navigation framework explained in \S\ref{sec:opinion_driven_framework}, with $u_i=10$.}
    \label{fig:2D_heatmap}
    \vspace{-1.5em}
\end{figure}

\subsubsection{Nonlinear Opinion Dynamics (NOD) Model} \label{nonlinear_opinion_dynamics_model}

The opinion state $z_i (t) = (z_{i1} (t), \cdots, z_{iN_s} (t))\in \mathbb{R} ^{N_s}$ represents robot~$i$'s preference for the available navigation strategies. This opinion state $z_i (t)$ is updated using the NOD model \cite{nonlinear_opinion_dynamics, tuning_reciprocity}, which is a continuous-time model that guides robots' decision-making processes. In the context of our navigation problem, the NOD model is particularly advantageous as it effectively resolves deadlocks, enabling multiple robots to coordinate their movements and successfully reach their destinations. According to the model, each robot $i$ maintains an opinion $z_{ij}(t)$ about each strategy $j$ and updates its opinion state dynamically based on the opinion states of others, governed by the following equation:
\begin{align} \label{eq:opinion_dynamics_continous}
\dot{z}_{i j} (t) = -d_i \big(z_{i j} (t) - u_i \textstyle\sum_{\substack{k=1 \\ k \neq i}}^{N_r} R ( A_{i k}^j \bar z_{k j} (t) ) - b_{ij} \big).
\end{align}

In \eqref{eq:opinion_dynamics_continous}, the variable $\bar z_{kj} (t)$, defined as 
\begin{align} \label{eq:relative_opinion}
\bar z_{kj} (t)= z_{kj} (t) - N_s^{-1} \textstyle \sum_{l=1}^{N_s} z_{kl} (t),
\end{align}
represents agent $k$'s relative opinion of strategy $j$ compared to other strategies. The magnitude and sign of $\bar z_{kj} (t)$ indicate the strength of robot~$k$'s preference or aversion for strategy~$j$. The function $R: \mathbb R \to [0, 1]$ is an increasing saturation function that maps robot~$k$'s opinion about strategy~$j$ to a normalized score within the range $[0,1]$. An example of such a function is given in \eqref{eq:R_hyperbolic_tangent}, which is used in simulations. The \textit{social term} $R ( A_{i k}^j \bar z_{k j} (t) )$ defines an incentive mechanism that encourages each robot~$i$ to coordinate with others. Here, $A_{ik}^j \in \{0, 1\}$ is a weighting factor used by robot~$i$ to evaluate strategy~$j$ based on its observation of robot~$k$'s opinion of the same strategy. For instance, if robot~$i$ cannot observe or estimate robot~$k$'s opinion state regarding strategy~$j$, the weight can be set to $A_{ik}^j = 0$; otherwise, we set $A_{ik}^j = 1$. The parameter $u_i$ quantifies the extent to which robot~$i$ considers the strategy preferences of other robots; a higher $u_i$ value indicates that robot~$i$ places greater importance on social interactions. The parameter $b_{ij}$ represents a bias term that captures robot~$i$'s predisposition toward strategy~$j$, where $b_{ij}>0$ indicates a preference, $b_{ij} = 0$ indicates neutrality, and $b_{ij} < 0$ indicates aversion. Finally, $d_i$ is a positive parameter that represents the time constant of robot~$i$, which governs its resistance to changes in its opinion state.

Robot~$i$ selects a strategy probabilistically, with the likelihood of selection being proportional to elements of $z_i (t)$. In particular, robot~$i$ selects strategy~$j$ with probability $x_{ij} (t)$ computed as 
\begin{equation} \label{eq:strategy_selection_softmax}
x_{i j} (t) = \frac{\exp ( \eta^{-1} z_{i j} (t) )}{\sum_{l=1}^{N_s} \exp (\eta^{-1} z_{i l} (t))},
\end{equation}
where $\eta$ is a positive constant, representing the level of randomness in the strategy selection.

\begin{remark}
The primary advantage of \eqref{eq:opinion_dynamics_continous} lies in its theoretical performance guarantee. For example, the main results in \cite{nonlinear_opinion_dynamics} show that consensus among agents' opinion states emerges as a \textit{stable equilibrium} in \eqref{eq:opinion_dynamics_continous}. Furthermore, \cite{bizyaeva2024multitopicbeliefformationbifurcations} explains how multiple such stable equilibria can arise, depending on the choice of model parameters in a generalized version of \eqref{eq:opinion_dynamics_continous}, which we utilize in \S\ref{sec:game_reduction}.
\end{remark}

With the parameter settings, $d_i=1$, $A_{ik}^j = 1$, $b_{ij} = 0$, and $\eta = 1$, Figs.~\ref{fig:2D_heatmap}a,b illustrate the steady-state strategy selections for a two-robot scenario ($N_r = 2$), starting with different initial preferences for strategy selection. The results are presented for two different values of $u_i$, highlighting how variations in the attention parameter $u_i$ influence the strategy selection outcomes.


\subsubsection{Strategy Estimation for Social Interaction} \label{strategy_perception_mechanism}
To implement the NOD model \eqref{eq:opinion_dynamics_continous}, each robot must evaluate the social term  $R(A_{ik}^j \bar z_{kj} (t) )$, which depends on the opinion states of other robots. However, in the absence of explicit communication, robots lack direct access to these opinion states or strategy selections. Instead, as robots update their strategy selections, they infer the opinions of others by observing their positions and velocities. This indirect approach allows each robot to gather the necessary information to update its opinion state, facilitating the possibility of achieving consensus on strategy selection.

Specifically, each robot observes the past path of others and compares them with the paths generated by the multi-robot path planner for all available strategies. For each strategy $j$ in $\{1, \cdots, N_s \}$, the $L_2$ norm quantifies the difference between the observed path and the path associated with strategy~$j$. The deviation is then mapped to an estimation of the social term as $r_{ik}^j (t) = R (A_{ik}^j \bar z_{kj}' (t))$, where\footnote{Note that $z_{kj}'(t) \leq 0$, and therefore it differs from the opinion $z_{kj} (t)$. However, after normalization and the application of the saturation function $R$, we interpret $r_{ik}^j (t)$ as an estimate of $R(A_{ik}^j \bar z_{kj} (t) )$.} 
\begin{subequations} \label{eq:est_opinion}
\begin{align}
z_{kj}'(t) &= - \textstyle K_1 \int_{t-T}^t \big\| X_{k}(\tau) - X_{k}^{(i,j)}(\tau) \big\|_2^2 \, \mathrm d\tau \label{eq:est_opinion_a} \\
\bar{z}_{kj}'(t) &= z_{kj}'(t) - N_s^{-1} \textstyle\sum_{l=1}^{N_s} z_{kl}'(t), \label{eq:est_opinion_b}
\end{align}
\end{subequations}
where $K_1$ is a scaling factor, set to $K_1 = 5$ in our simulations.

In \eqref{eq:est_opinion}, the term $X_{k}(\tau)$ represents the state of robot $k$, while $X_{k}^{(i,j)}(\tau)$ denotes the state of robot $k$ inferred by robot $i$ under strategy $j$, as determined by the multi-robot path planner. 
The parameter $T$ defines the time window over which the state trajectory of robot~$k$ is used to estimate the social term. Using \eqref{eq:est_opinion}, a larger discrepancy between the path inferred by a strategy and the robot's actual traveled path corresponds to a lower opinion value for the evaluated strategy. By employing this estimation scheme, each robot infers the opinion states of others and incorporates this information to update its own opinion state using the NOD model \eqref{eq:opinion_dynamics_continous}, substituting $r_{ik}^j(t)$ for $R(A_{ik}^j \bar z_{kj} (t) )$.

To examine the effect of the social term estimation \eqref{eq:est_opinion} on the robots' strategy selections, we employ the opinion-driven navigation framework -- which integrates \eqref{eq:est_opinion} with a multi-robot path planner, as detailed in \S\ref{sec:opinion_driven_framework} -- using the parameter settings $d_i=1$, $A_{ik}^j = 1$, $b_{ij} = 0$, $\eta = 1$, and $u_i = 10$. Fig.~\ref{fig:2D_heatmap_c} presents the resulting strategy selections of two robots, averaged over multiple simulations of the \textit{2-Robot Case-2} scenario shown in Fig.~\ref{fig:configuration_table}. Compared to the heatmap in Fig.~\ref{fig:2D_heatmap_b}, which is generated directly from \eqref{eq:opinion_dynamics_continous} and \eqref{eq:strategy_selection_softmax}, the overall trend of the strategy selections based on initial opinion states remains consistent. Notably, when the robots have similarly strong but opposing preferences for the two strategies, they tend to alternate in passing through the corridor first. Consequently, the average strategy selection converges toward an approximately equal likelihood between the two options.

\section{Game Reduction for Scalable Interaction} \label{sec:game_reduction}

A notable limitation of the framework described in \S\ref{approach} is that each robot must compute the paths of all other robots to estimate the social term in \eqref{eq:opinion_dynamics_continous}. This requirement significantly increases computational complexity, as the search space for multi-robot path planning grows exponentially with the number of robots. To address this issue, we propose a technique that reduces the computational burden by selecting a smaller subset of robots for the path planning. Each robot then updates its opinion state based on the paths of the robots within this subset. We refer to this method as the \textit{game reduction} technique, as it effectively reduces the number of players (other robots) each robot considers in the navigation game.

To apply the game reduction technique, we adopt a generalized version of \eqref{eq:opinion_dynamics_continous}, as proposed in \cite{bizyaeva2024multitopicbeliefformationbifurcations}, where each agent~$i$ interacts only with a subset $\mathcal N_i (t) \subset \{1, \cdots, N_r\} \setminus \{i\}$ of robots, referred to as its \textit{game players}:
\begin{equation} \label{eq:opinion_dynamics_continous_loca_interaction}
\dot{z}_{i j}(t) \!=\! -d_i \Big(z_{i j} (t) - u_i \textstyle\sum_{k \in \mathcal N_i (t)} \sum_{l=1}^{N_s} R ( A_{i k}^{jl} \bar z_{k l} (t)) - b_{ij} \Big),
\end{equation}
where $\mathcal N_i (t)$ represents the game players of robot~$i$, whose opinion states can be estimated using the method outlined in \S\ref{strategy_perception_mechanism}. Note that $\mathcal N_i (t)$ is time-dependent, allowing the game players of each robot to vary over time. When $\mathcal N_i(t)$ includes all other robots and $A_{ik}^{jl}$ is defined as $A_{ik}^{jl} = A_{ik}^{j}$ if $j = l$, and $A_{ik}^{jl} = 0$ otherwise, it follows that \eqref{eq:opinion_dynamics_continous_loca_interaction} generalizes  \eqref{eq:opinion_dynamics_continous}.

\subsection{Game Player Selection} \label{sec:game_player_selection}
We describe a procedure for selecting game players $\mathcal N_i(t)$ using a \textit{conflict likelihood metric}, defined as follows. Each robot~$i$ assesses the likelihood of conflict with others by considering two key factors: 1) their proximity to the entrance of the narrow corridor, prioritizing robots that are close, particularly when robot~$i$ is also nearby, and 2) their proximity to robot~$i$, focusing on those in its immediate vicinity when multiple such robots are present. 

Based on these factors, we define the conflict likelihood metric between two robots, $i$ and $j$, as follows:
\begin{equation} \label{conflict_score}
    L^\text{conflict}_{ij} = \frac{1}{d_{ij} + \delta} + \frac{1}{d_{i, \text{corridor}} + d_{j, \text{corridor}} + \delta},
\end{equation}
where $\delta$ is a positive constant ($\delta=1$ in our simulation) introduced to ensure that \eqref{conflict_score} is well-defined for all nonnegative values of $d_{ij}$, $d_{i,\text{corridor}}$, and $d_{j, \text{corridor}}$. Additionally, $\delta$ serves to impose an upper bound on $L^\text{conflict}_{ij}$. The variable $d_{ij}$ represents the Euclidean distance between robots $i$ and $j$, while $d_{i,\text{corridor}}$ and $d_{j, \text{corridor}}$ denote the Euclidean distance of robots $i$ and $j$ to the entrance of the corridor, respectively. In our scenario, as illustrated in Fig.~\ref{fig:env_pic}, if a robot approaches the corridor from the left side, the entrance point is set to $(-3, 0)$, and if it approaches from the right side, the entrance point is $(3, 0)$. Once a robot passes the entrance point, its distance to the corridor is considered zero until it exits. 

Suppose each robot can have $k (<N_r)$ game players. At time $t$, each robot~$i$ evaluates \eqref{conflict_score} and selects $k$ robots based on the highest conflict likelihoods from all possible subsets of size~$k$, maximizing:
\begin{align} \label{eq:game_player_selection}
\textstyle \max_{|\mathcal{N}_i (t)|=k} \sum_{j \in \mathcal N_i (t)} L^\text{conflict}_{ij}.
\end{align}
Consequently, the selected set $\mathcal N_i (t)$ consists of the robots most likely to pose a potential conflict with robot~$i$. 

\subsection{Opinion State Update} \label{sec:opinion_state_update}
Similar to the social term estimation explained in \S\ref{strategy_perception_mechanism}, implementing \eqref{eq:opinion_dynamics_continous_loca_interaction} involves computing $r_{ik}^{jl} (t) = R (A_{ik}^{jl} \bar z_{kl}' (t))$ and substituting it for $R (A_{ik}^{jl} \bar z_{kl} (t))$, where $\bar z_{kl}'(t)$ is determined by \eqref{eq:est_opinion}. To explain how $A_{ik}^{jl}$ is selected, recall that each strategy dictates the order in which $N_r$ robots pass through the corridor, while each robot~$i$ estimates only the paths of its game players $\mathcal N_i(t)$ under the game reduction technique. We set $A_{ik}^{jl} = 1$ only when robot~$k$ is one of robot~$i$'s game players -- that is, $k \in \mathcal N_i(t)$ -- and strategies~$j$ and $l$ prescribe the same passing order for robot~$i$ and its game players $\mathcal N_i (t)$. Otherwise, $A_{ik}^{jl}$ is set to $0$. This selection encourages each robot to avoid deadlocks with its game players by prioritizing strategies that preserve a consistent passing order with its game players -- robots that are most likely to experience conflicts with robot~$i$.

\section{Evaluation} \label{simulation_results}

\subsection{NOD Model, Multi-Robot Path Planner, and Motion Controller Implementation}
\label{sec:opinion_driven_framework}
\subsubsection{NOD Model} \label{sec:nod_model_simulation}
Each robot~$i$ updates $r_{ik}^{jl} (t)$ at discrete time steps $t \in \{T, 2T, \cdots \}$ using the path data of robot~$k$ over the interval $[t-T, t]$ and use it to update the opinion state. To achieve this, we employ a time-discretized version of \eqref{eq:opinion_dynamics_continous_loca_interaction}. Based on the stochastic approximation method for a similar model used in \cite[\S VII]{9022871}, the discrete-time approximation is given by:
\begin{multline} \label{eq:opinion_dynamics_discrete}
    z_{i j}(t+T) = z_{i j}(t) \\ - \alpha(t) d_i \Big( z_{i j}(t) - u_i \textstyle \sum_{k \in \mathcal N_i (t)} \sum_{l=1}^{N_s} r_{ik}^{jl}(t) - b_{ij} \Big)
\end{multline}
for $t \in \{0, T, 2T, \cdots \}$, where the step size $\alpha(t)$ satisfies
\begin{align}
    \textstyle \sum_{t=0}^\infty \alpha(t) = \infty ~~\text{and}~~ \sum_{t=0}^\infty \alpha^2(t) < \infty.
\end{align}
This discrete-time formulation ensures numerical stability while preserving the key dynamics of the original continuous-time model \eqref{eq:opinion_dynamics_continous_loca_interaction}. 
In our simulations, we define the saturation function $R$ using the hyperbolic tangent function as follows:
\begin{align} \label{eq:R_hyperbolic_tangent}
    r_{ik}^{jl}(t) = \textstyle R(A_{ik}^{jl} \bar z_{kl}'(t)) = \frac{1}{2} (\tanh(A_{ik}^{jl} \bar z_{kl}'(t)) + 1).
\end{align}
Additionally, for all robots, we use the following fixed parameters $d_i = 1$, $u_i = 100$, and $T = 1$ for \eqref{eq:opinion_dynamics_discrete}, $\eta = 1$ for \eqref{eq:strategy_selection_softmax},\footnote{The parameter selection is guided by the analysis in \cite{nonlinear_opinion_dynamics} and \cite{bizyaeva2024multitopicbeliefformationbifurcations}, as well as by a trial-and-error process, which show that when robots are homogeneous and have identical parameters in the NOD model, they can reach a consensus on strategy selection given a sufficiently large $u_i$. Also, both the theoretical analysis in these references and additional simulation results (not reported here) confirm that small variations in these parameters do not significantly affect the simulation outcomes.} and the step size defined as $\alpha(t) = \frac{1}{t+1}$.\footnote{In the actual implementation, we used $\alpha(t) = \max(\frac{1}{t+1}, h)$, where $h = 0.05$ is a predefined lower bound. This ensures that $\alpha(t)$ remains bounded away from zero during simulations over finite time intervals by enforcing a minimum step size $h$.} 

\subsubsection{Multi-Robot Path Planner and Motion Controller} \label{sec:multirobot_path_planner}
The primary objective of the multi-robot path planner is to compute paths for multiple robots from their current positions to specified destinations. As described in \S\ref{strategy_perception_mechanism} and \S\ref{sec:opinion_state_update}, this planner plays a crucial role in computing $r_{ik}^{jl} (t)$ in \eqref{eq:opinion_dynamics_discrete}. Since the specific choice of the planner has minimal impact on the overall performance of our framework, we use the Probabilistic Roadmap (PRM) Star algorithm \cite{samplingBasedAlgorithms_PRM} for simplicity. This algorithm is well-suited to our needs due to its ability to precompute PRMs, enabling efficient reuse for generating paths across various robot configurations and destinations.

For our simulation studies, we generate PRMs for $N_r = 2, 3, 4$, where one PRM is generated for each strategy, resulting in a total of $N_r!$ PRMs for each fixed $N_r$. We use the PRM implementation provided by the OMPL library \cite{sucan2012the-open-motion-planning-library}. To ensure sufficient roadmap density, the PRMs are computed with extended generation times depending on the number of robots: $10$ minutes for $N_r=1$, $30$ minutes for $N_r=2$, $60$ minutes for $N_r=3$, and $120$ minutes for $N_r=4$. These precomputed PRMs are then used in all subsequent simulations, with all robots sharing the same set of PRMs.

Once the precomputed PRMs generate paths for multiple robots to their respective destinations, Model Predictive Control (MPC) is employed as a motion control algorithm to ensure the robots to follow their PRM paths while avoiding collisions with obstacles. Specifically, we use a B-spline representation for the MPC implementation, with a cost function designed to optimize a B-spline path that closely follows the PRM path while avoiding collisions with other robots' predicted PRM paths and obstacles in the environment. When the game reduction technique described in \S\ref{sec:game_reduction} is applied, the precomputed PRMs do not generate paths for non-game players, i.e., other robots outside the set $\mathcal{N}_i (t)$. In such cases, robot~$i$ treats these robots as static obstacles and uses MPC to avoid collision with them while navigating along its assigned PRM path. 
The implementation details of PRM and MPC are provided in Appendix.

\subsubsection{Integration}

\begin{algorithm} [t]
\footnotesize
\caption{Opinion-Driven Navigation for Robot~$i$} \label{alg:framework}
    \DontPrintSemicolon
    \SetKwFunction{FMultiRobotPathPlanning}{MultiRobotPathPlanning}
    \SetKwProg{Fn}{Function}{:}{}
    \SetKwFunction{FOpinionStateUpdate}{OpinionStateUpdate}
    \SetKwProg{Fn}{Function}{:}{}
    \SetKwFunction{proc}{RobotNavigation}
    \SetKwProg{myproc}{Procedure}{:}{}

    \Fn{\FOpinionStateUpdate{$X_{t-T:t}^{(i,1)}, \cdots, X_{t-T:t}^{(i,N_s)}$}} {

        Using \eqref{eq:est_opinion} and \eqref{eq:opinion_dynamics_discrete}, along with $(X_{t-T:t}^{(i,1)}, \cdots, X_{t-T:t}^{(i,N_s)})$, update the opinion state $z_{i}(t) = (z_{i1}(t), \cdots, z_{i N_s}(t))$ of robot~$i$.
        
        \Return $z_{i}(t)$ 
    }

    \Fn{\FMultiRobotPathPlanning{$j$}} {
        Compute paths $X_{t:T_f}^{(i,j)} = (X^{(i,j)}(t), \cdots, X^{(i,j)}(T_f))$ of multiple robots, including robot $i$ and those in $\mathcal N_i(t)$, from their current state $X (t)$ to their respective destinations under navigation strategy~$j$.
        
        \Return $X_{t:T_f}^{(i,j)}$    
    }
    
    \myproc{\proc{$t$}} {

    Retrieve previous multi-robot paths $(X_{t-T:t}^{(i,1)}, \cdots, X_{t-T:t}^{(i,N_s)})$.

    $z_i(t)$ = \FOpinionStateUpdate($X_{t-T:t}^{(i,1)}, \cdots, X_{t-T:t}^{(i,N_s)}$)

    Using \eqref{eq:game_player_selection}, update the game player set $\mathcal N_i(t)$.

    \For{$j = 1, \cdots, N_s$} {
        Compute $x_{ij}(t)$ using \eqref{eq:strategy_selection_softmax}.
    
        $X_{t:T_f}^{(i,j)}$ =  \FMultiRobotPathPlanning($j$)
    }

    Select strategy $S_i$ according to the probability $(x_{i 1}(t), \cdots, x_{i N_s}(t))$.

    Control robot~$i$ to navigate along $(X_i^{(i,S_i)}(t), \cdots, X_i^{(i,S_i)} (T_f))$.
    }
\end{algorithm}

We integrate the time-discretized NOD model, precomputed PRMs, and MPC to develop the proposed opinion-driven navigation framework, as outlined in Algorithm~\ref{alg:framework}.
In this framework, at discrete time steps $t = 0, T, 2T, \cdots$, each robot updates its opinion state using \eqref{eq:opinion_dynamics_discrete}, updates its game players $\mathcal N_i (t)$ according to \eqref{eq:game_player_selection}, and selects a new navigation strategy using \eqref{eq:strategy_selection_softmax}. 


\subsection{Simulation Results}

\begin{figure}
    \centering
        \includegraphics[width=0.45\textwidth]{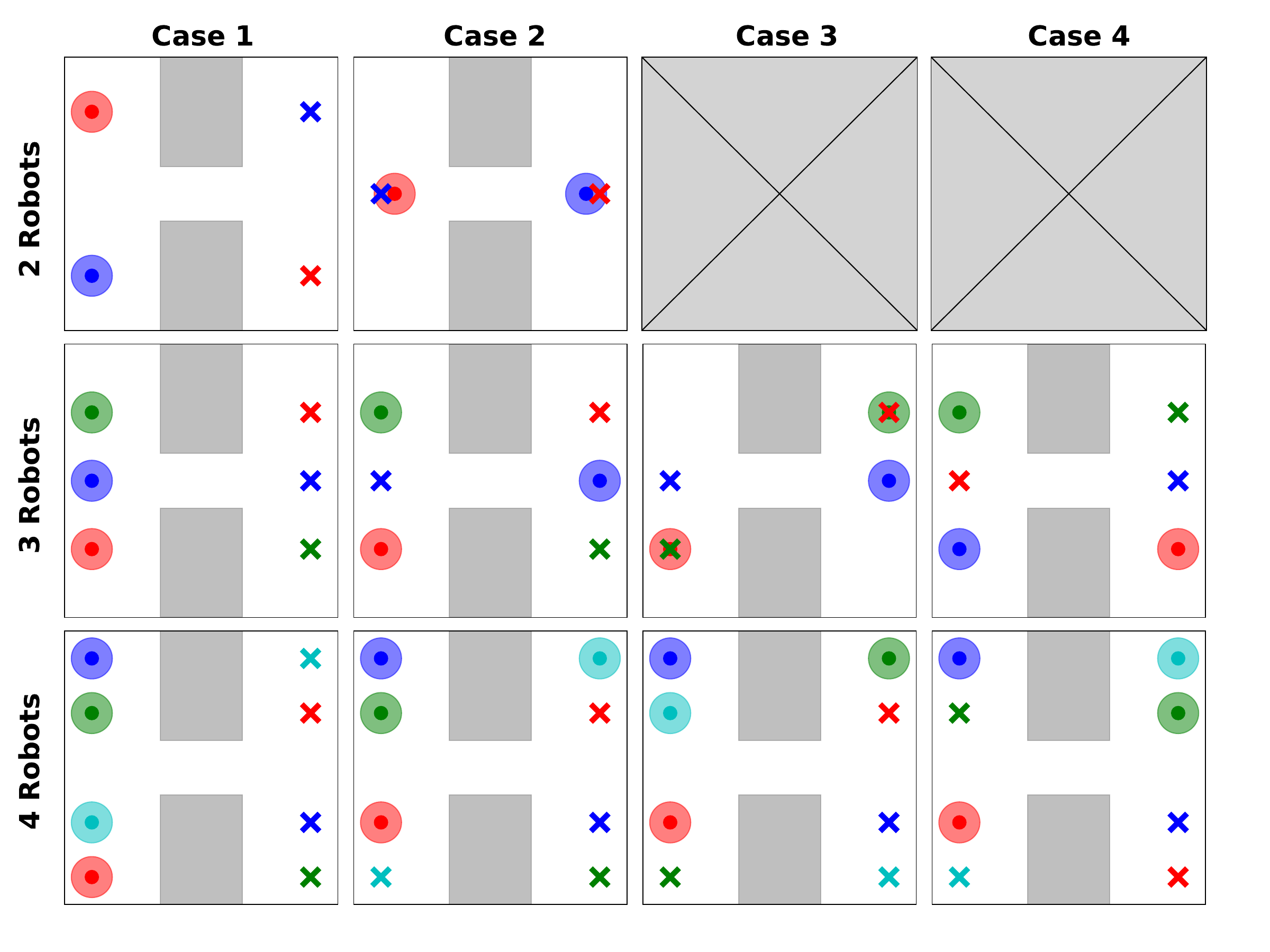}
    \caption{Illustrations of various evaluation scenarios. The colored circles indicate the robots’ origins, while the X-marks in the same colors represent their corresponding destinations.}
    \label{fig:configuration_table}
\end{figure}

\begin{figure*}[t]
    \centering
    \begin{subfigure}[b]{\textwidth}
    \centering
    \includegraphics[width=0.82\textwidth]{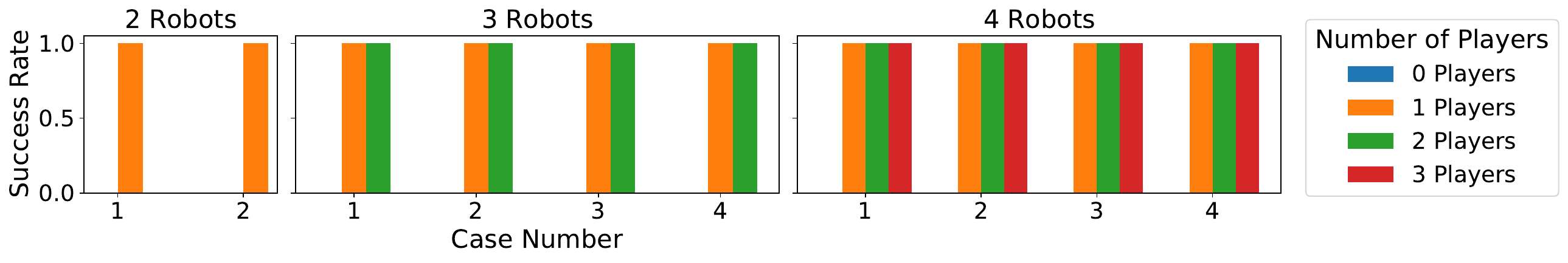}
    \end{subfigure}
    \caption{Success rates of corridor navigation across different scenarios, each case is illustrated in Fig.~\ref{fig:configuration_table}, and for varying numbers $|\mathcal N_i(t)|$ of game players. The success rate represents the percentage of simulation trials in which all robots successfully navigated through the corridor and reached their destinations. }
    \label{fig:success_rates}
    \vspace{1.0em}
    
    \centering
    \begin{subfigure}[b]{0.22\textwidth}
        \centering
        \includegraphics[page=10, width=0.8\textwidth]{Basic/3r_3p_u100_sim_38080790_82_timestep.pdf}
    \end{subfigure}
    \begin{subfigure}[b]{0.22\textwidth}
        \centering
        \includegraphics[page=60, width=0.8\textwidth]{Basic/3r_3p_u100_sim_38080790_82_timestep.pdf}
    \end{subfigure}
    \begin{subfigure}[b]{0.22\textwidth}
        \centering
        \includegraphics[page=130, width=0.8\textwidth]{Basic/3r_3p_u100_sim_38080790_82_timestep.pdf}
    \end{subfigure}
    \begin{subfigure}[b]{0.22\textwidth}
        \centering
        \includegraphics[page=180, width=0.8\textwidth]{Basic/3r_3p_u100_sim_38080790_82_timestep.pdf}
    \end{subfigure}

    \begin{subfigure}[b]{0.22\textwidth}
        \centering
        \includegraphics[page=10, width=\textwidth]{Basic/3r_3p_u100_sim_38080790_82_actions_timestep.pdf}
        \caption{}
        \label{fig:subfig1}
    \end{subfigure}
    \begin{subfigure}[b]{0.22\textwidth}
        \centering
        \includegraphics[page=60, width=\textwidth]{Basic/3r_3p_u100_sim_38080790_82_actions_timestep.pdf}
        \caption{}
        \label{fig:subfig2}
    \end{subfigure}
    \begin{subfigure}[b]{0.22\textwidth}
        \includegraphics[page=130, width=\textwidth]{Basic/3r_3p_u100_sim_38080790_82_actions_timestep.pdf}
        \caption{}
        \label{fig:subfig3}
    \end{subfigure}
    \begin{subfigure}[b]{0.22\textwidth}
        \includegraphics[page=180, width=\textwidth]{Basic/3r_3p_u100_sim_38080790_82_actions_timestep.pdf}
        \caption{}
        \label{fig:subfig3}
    \end{subfigure}
    \caption{Snapshots from the simulation in \textit{3-Robot Case-4}, where each robot selects all others as its game players. The top row illustrates how the robots coordinate to reach their destinations, while the bottom row shows the evolution of their strategy selections over time. The legend in each bottom-row figure indicates the current strategy selected by each robot.
    }
    \label{fig:basic_3robots_3players_case4}
    \vspace{1.0em}
    
    \centering
    \begin{subfigure}[b]{0.19\textwidth}
        \includegraphics[page=20, width=\textwidth]{Basic/4r_2p_u100_sim_38087481_393_timestep.pdf}
    \end{subfigure}
    \begin{subfigure}[b]{0.19\textwidth}
        \includegraphics[page=100, width=\textwidth]{Basic/4r_2p_u100_sim_38087481_393_timestep.pdf}
    \end{subfigure}
    \begin{subfigure}[b]{0.19\textwidth}
        \includegraphics[page=200, width=\textwidth]{Basic/4r_2p_u100_sim_38087481_393_timestep.pdf}
    \end{subfigure}
    \begin{subfigure}[b]{0.19\textwidth}
        \includegraphics[page=250, width=\textwidth]{Basic/4r_2p_u100_sim_38087481_393_timestep.pdf}
    \end{subfigure}
    \begin{subfigure}[b]{0.19\textwidth}
        \includegraphics[page=280, width=\textwidth]{Basic/4r_2p_u100_sim_38087481_393_timestep.pdf}
    \end{subfigure}
    
    \caption{Snapshots from the simulation for \textit{4-Robot Case-4}, where each robot is allowed to select only one game player. The arrows and their colors indicate the selected game player for each robot. Note that these selections are dynamically updated over time based on \eqref{eq:game_player_selection}.
    }
    \label{fig:game_player_selection}
    \vspace{-1.0em}
\end{figure*}

We conducted extensive simulation studies involving 2, 3, and 4 robots, each starting from different origins and moving toward different destinations, to evaluate the performance of our framework across a range of scenarios. Fig.~\ref{fig:configuration_table} illustrates the evaluated scenarios.\footnote{We focus on scenarios with up to $4$ robots, as the confined environment leads to congestion when simulating more than $4$ robots. This congestion prevents the MPC from enabling them to pass each other, even though they agree on a single strategy.} First, we examined scenarios without any initial preference for a specific strategy and no biases ($z_{ij}(0) = 0, b_{ij} = 0, ~ \forall (i,j) \in \{1, \cdots, N_r\} \times \{1, \cdots, N_s\}$), and all robots are interacting with one another ($\mathcal N_i (t) = \{1, \cdots, N_r\} \setminus \{i\}$). Fig.~\ref{fig:success_rates} summarizes the evaluation results from multiple trials, with at least 40 trials conducted for each scenario. 

Fig.~\ref{fig:basic_3robots_3players_case4} illustrates one particular instance of the \textit{3-Robot Case-4} scenario. As shown in the figure, by estimating the intentions behind each other's movements and using the NOD model, the robots successfully reach a consensus on strategy selection, allowing them to navigate the corridor successfully. In this particular simulation, robot~$3$ (green) decides to pass through the narrow corridor first, followed by robot~$2$ (blue), and finally robot~$1$ (red). This sequence occurs because, while robot~$1$ is waiting for robot~$3$ to proceed, robot~$2$ recognizes that robot~$1$ is also waiting for it to pass first. Consequently, robot~$2$ decides to proceed before robot~$1$, and robot~$1$ waits for robot~$2$ to pass before moving forward. 

\begin{table}
\centering
\caption{Evaluation results for the \textit{3-Robot Case-2} under distance-based initial opinion states.}
\label{tab:conditions_effect_on_strategy_selection}
\resizebox{\columnwidth}{!}{
\begin{tabular}{|c|c|c|c|c|c|c|c|}
\hline
\textbf{\boldmath $z_{ij}(0)$} & {\boldmath $S_1$} & {\boldmath $S_2$} & {\boldmath $S_3$} & {\boldmath $S_4$} & {\boldmath $S_5$} & {\boldmath $S_6$} \\
\hline \hline
0 & 10.0\%  & 21.67\%  & 10.00\%  & 18.33\% & 6.67\%  & 33.33\% \\ 
\hline
Eq.~\eqref{eq:distance_opinion} & 0.0\%  & 0.0\%  & 40.0\%  & 60.0\% & 0.0\%  & 0.0\% \\ 
\hline
\end{tabular}
}
\vspace{.5em}
\centering
\caption{Evaluation results for the \textit{3-Robot Case-3} under different bias.}
\label{tab:bias_effect_on_strategy_selection}
\resizebox{\columnwidth}{!}{
\begin{tabular}{|c|c|c|c|c|c|c|c|}
\hline
\textbf{Bias} & {\boldmath $S_1$} & {\boldmath $S_2$} & {\boldmath $S_3$} & {\boldmath $S_4$} & {\boldmath $S_5$} & {\boldmath $S_6$} \\ 
\hline \hline             
0 & 6.67\% & 41.67\%  & 8.33\% & 23.33\%  & 1.67\% & 18.33\% \\
\hline
150 & 0.0\%  & 0.0\%  & 53.33\%  & 0.0\% & 46.67\%  & 0.0\% \\ 
\hline
\end{tabular}
}

\end{table}

\subsubsection{Effect of the Initial Condition} 
Next, we conducted a set of 30 simulations in which the initial strategy preference $z_{ij} (0)$ was determined based on the robots’ proximity to the entrance of the corridor. The objective of these simulations was to evaluate whether our framework enables robots to reach a consensus on prioritizing passage for those positioned closer to the corridor. To achieve this, $z_{ij} (0)$ is defined based on the distances between the robots’ origins and the entrance of the corridor, where the distances are measured using the same metric as described in \S\ref{sec:game_player_selection}. 

In particular, suppose strategy~$j$ specifies that the robots pass in the order $(i_1, \cdots, i_{N_r})$. We define the likelihood that robot~$i_l$ proceeds immediately after robots $i_1, \cdots, i_{l-1}$ as:
\begin{align}\label{eq:order_likelihood}
L ( i_l \,|\, i_1, \cdots, i_{l-1}) = \frac{\exp (-d_{i_l,\text{corridor}}) }{\sum_{k=l}^{N_r}  \exp (-d_{i_k,\text{corridor}})},
\end{align}
where $d_{i_l, \text{corridor}}$ denotes the initial distance of robot~$i_l$ from the entrance of the corridor. According to this definition, if robot~$i_l$ positioned closer to the entrance than the remaining robots $i_{l+1}, \cdots, i_{N_r}$, it is assigned a higher likelihood of passing earlier.
Then, we calculate $z_{ij} (0)$ as:
\begin{align}\label{eq:distance_opinion}
    z_{ij} (0) 
    &=  \Pi_{l=1}^{N_r-1}  K_2 \cdot L ( i_l \,|\, i_1, \cdots, i_{l-1}),
\end{align}
where $K_2$ is a scaling factor, set to $K_2 = 10$ for the simulations. Consequently, $z_{ij} (0)$ can be interpreted as the likelihood of $N_r$ robots passing in the order $(i_1, \cdots, i_{N_r})$. By assigning the initial opinion state in this manner, the robots are effectively guided to prioritize the passing order based on their initial positions.

Table~\ref{tab:conditions_effect_on_strategy_selection} reports the effect of initial opinion states, as defined by \eqref{eq:distance_opinion}, on the strategy selection in the \textit{3-Robot Case-2} scenario, illustrated in Fig.~\ref{fig:configuration_table}. Notably, robot 2 (blue) on the right is the closest to the corridor, while robot 1 (red) and 3 (green) on the left are positioned at an equal distance from the corridor. Initially, multiple simulation runs were conducted without any preference for a specific strategy ($z_{ij}(0) = 0, ~ \forall (i,j) \in \{1, \cdots, N_r\} \times \{1, \cdots, N_s\}$). As shown in the first row of Table~\ref{tab:conditions_effect_on_strategy_selection}, 
in this condition, all strategies were selected with minor variations in frequency. However, when the initial opinion $z_{ij}(0)$ was determined using \eqref{eq:distance_opinion}, as reported in the second row of Table~\ref{tab:conditions_effect_on_strategy_selection}, only strategy 3: (2, 1, 3) and strategy 4: (2, 3, 1) were selected, both of which prioritize robot~2 (blue) passing first. These results demonstrate that initializing the opinion state based on proximity to the corridor effectively influences the robots' strategy selection. 

\subsubsection{Effect of the Bias}

We examined how the bias affects the strategy selections. To this end, in the \textit{3-Robot Case-3} scenario, illustrated in Fig.~\ref{fig:configuration_table}, we assigned the bias $b_{ij}$ as follows: $b_{ij} = 150$ if $j \in \{3, 5\}$, and $b_{ij} = 0$ otherwise. Note that strategies~$3$ and $5$ correspond to cases where robot~$1$ (red) passes second, positioned between robot~$2$ (blue) and robot~$3$ (green). Table~\ref{tab:bias_effect_on_strategy_selection} summarizes the effect of the bias on the strategy selection. In the absence of the bias, all strategies were selected with varying frequencies. However, when the NOD model was biased toward selecting strategy~$3$ or strategy~$5$, the robots consistently reached a consensus on one of these two preferred strategies. These results demonstrate that, by appropriately introducing bias, our framework can effectively avoid \textit{unfair} cases where robot~$1$ (red) is forced to wait for both other robots to pass first -- corresponding to strategies~$4$ and $6$ -- before it can proceed.

\subsubsection{Effectiveness of Game Reduction Technique}
We evaluate the effectiveness of the game reduction technique described in \S\ref{sec:game_reduction} by applying it to the same scenarios previously simulated without the reduction, as shown in Fig.~\ref{fig:configuration_table}. In this evaluation, game players are selected based on \eqref{eq:game_player_selection}, replacing the all-to-all interaction used in earlier simulations. Fig.~\ref{fig:success_rates} summarizes the simulation results, comparing the success rates obtained with the game reduction technique against those from all-to-all interaction scenarios. Additionally, Fig.~\ref{fig:game_player_selection} visualizes the \textit{4-Robot Case-4} scenario, where each robot is restricted to interacting with only one other robot at a time ($|\mathcal N_i(t)| = 1$). At every time step $t \in \{0, T, 2T, \cdots \}$, the robots re-evaluate their game player selection using the conflict likelihood \eqref{conflict_score}.

As shown in Fig.~\ref{fig:success_rates}, the success rate remains close to $100\%$ even when each robot interacts with only a subset of the others. This demonstrates that effective narrow corridor navigation can still be achieved with reduced interaction complexity. However, it is important to note that when robots navigate without considering any other robots as game players -- i.e., under $|\mathcal N_i(t)| = 0$ and all others are treated as static obstacles -- the success rate drops to $0\%$ due to deadlocks resulting from the lack of coordination.

\section{Conclusions and Future Plans} \label{conclusion}
This work investigates the multi-robot navigation problem in environments with a narrow corridor, where multiple robots must navigate through a constrained passage to reach their destinations. The primary challenge arises from the absence of explicit communication between robots and the lack of centralized coordination for managing the corridor passage. To address this challenge, we propose a multi-robot navigation framework that integrates the NOD model with the multi-robot path planning and motion control algorithms. 

For future research, we plan to validate our framework through experiments involving human participants, where factors such as inter-agent visibility, sensing error, and heterogeneity in decision-making may influence the perception and evaluation of social interactions. Although the framework already demonstrates strong performance using the game reduction technique, its effectiveness could be further enhanced by incorporating implicit communication mechanisms, such as signaling. Additionally, we aim to explore a data-driven approach for parameter selection. In our simulation studies, parameters were tuned manually through a trial-and-error process; however, automating this optimization would be valuable for applying the framework across diverse environments.

\bibliographystyle{IEEEtran} 
\bibliography{references}

\begin{thebibliography}{10}
\providecommand{\url}[1]{#1}
\csname url@samestyle\endcsname
\providecommand{\newblock}{\relax}
\providecommand{\bibinfo}[2]{#2}
\providecommand{\BIBentrySTDinterwordspacing}{\spaceskip=0pt\relax}
\providecommand{\BIBentryALTinterwordstretchfactor}{4}
\providecommand{\BIBentryALTinterwordspacing}{\spaceskip=\fontdimen2\font plus
\BIBentryALTinterwordstretchfactor\fontdimen3\font minus
  \fontdimen4\font\relax}
\providecommand{\BIBforeignlanguage}[2]{{%
\expandafter\ifx\csname l@#1\endcsname\relax
\typeout{** WARNING: IEEEtran.bst: No hyphenation pattern has been}%
\typeout{** loaded for the language `#1'. Using the pattern for}%
\typeout{** the default language instead.}%
\else
\language=\csname l@#1\endcsname
\fi
#2}}
\providecommand{\BIBdecl}{\relax}
\BIBdecl

\bibitem{nonlinear_opinion_dynamics}
A.~Bizyaeva, A.~Franci, and N.~E. Leonard, ``Nonlinear opinion dynamics with
  tunable sensitivity,'' \emph{IEEE Transactions on Automatic Control},
  vol.~68, no.~3, pp. 1415--1430, 2023.

\bibitem{bizyaeva2024multitopicbeliefformationbifurcations}
------, ``Multi-topic belief formation through bifurcations over signed social
  networks,'' \emph{IEEE Transactions on Automatic Control}, pp. 1--16, 2025.

\bibitem{tuning_reciprocity}
S.~Park, A.~Bizyaeva, M.~Kawakatsu, A.~Franci, and N.~E. Leonard, ``Tuning
  cooperative behavior in games with nonlinear opinion dynamics,'' \emph{IEEE
  Control Systems Letters}, vol.~6, pp. 2030--2035, 2021.

\bibitem{moller2021survey}
R.~M{\"o}ller, A.~Furnari, S.~Battiato, A.~H{\"a}rm{\"a}, and G.~M. Farinella,
  ``A survey on human-aware robot navigation,'' \emph{Robotics and Autonomous
  Systems}, vol. 145, p. 103837, 2021.

\bibitem{gao2022evaluation}
Y.~Gao and C.-M. Huang, ``Evaluation of socially-aware robot navigation,''
  \emph{Frontiers in Robotics and AI}, vol.~8, p. 420, 2022.

\bibitem{mavrogiannis2022social}
C.~Mavrogiannis, P.~Alves-Oliveira, W.~Thomason, and R.~A. Knepper, ``Social
  momentum: Design and evaluation of a framework for socially competent robot
  navigation,'' \emph{ACM Transactions on Human-Robot Interaction (THRI)},
  vol.~11, no.~2, pp. 1--37, 2022.

\bibitem{mavrogiannis2021core}
C.~Mavrogiannis, F.~Baldini, A.~Wang, D.~Zhao, P.~Trautman, A.~Steinfeld, and
  J.~Oh, ``Core challenges of social robot navigation: A survey,''
  \emph{arXiv:2103.05668}, 2021.

\bibitem{mirsky2021prevention}
R.~Mirsky, X.~Xiao, J.~Hart, and P.~Stone, ``Prevention and resolution of
  conflicts in social navigation--a survey,'' \emph{arXiv:2106.12113}, 2021.

\bibitem{mirsky2021conflict}
------, ``Conflict avoidance in social navigation--a survey,''
  \emph{arXiv:2106.12113}, 2021.

\bibitem{burgoon1976toward}
J.~K. Burgoon and S.~B. Jones, ``Toward a theory of personal space expectations
  and their violations,'' \emph{Human communication research}, vol.~2, no.~2,
  pp. 131--146, 1976.

\bibitem{daza2021approach}
M.~Daza, D.~Barrios-Aranibar, J.~Diaz-Amado, Y.~Cardinale, and J.~Vilasboas,
  ``An approach of social navigation based on proxemics for crowded
  environments of humans and robots,'' \emph{Micromachines}, vol.~12, no.~2, p.
  193, 2021.

\bibitem{kollmitz2020learning}
M.~Kollmitz, T.~Koller, J.~Boedecker, and W.~Burgard, ``Learning human-aware
  robot navigation from physical interaction via inverse reinforcement
  learning,'' in \emph{2020 IEEE/RSJ International Conference on Intelligent
  Robots and Systems (IROS)}, 2020, pp. 11\,025--11\,031.

\bibitem{baghi2021sample}
B.~H. Baghi and G.~Dudek, ``Sample efficient social navigation using inverse
  reinforcement learning,'' \emph{arXiv:2106.10318}, 2021.

\bibitem{okal2016learning}
B.~Okal and K.~O. Arras, ``Learning socially normative robot navigation
  behaviors with bayesian inverse reinforcement learning,'' in \emph{2016 IEEE
  International Conference on Robotics and Automation (ICRA)}, 2016, pp.
  2889--2895.

\bibitem{baghi2022sesno}
B.~H. Baghi, A.~Konar, F.~Hogan, M.~Jenkin, and G.~Dudek, ``Sesno: Sample
  efficient social navigation from observation,'' in \emph{2022 IEEE/RSJ
  International Conference on Intelligent Robots and Systems (IROS)}.\hskip 1em
  plus 0.5em minus 0.4em\relax IEEE, 2022, pp. 9164--9171.

\bibitem{bera2019emotionally}
A.~Bera, T.~Randhavane, R.~Prinja, K.~Kapsaskis, A.~Wang, K.~Gray, and
  D.~Manocha, ``The emotionally intelligent robot: Improving social navigation
  in crowded environments,'' \emph{arXiv:1903.03217}, 2019.

\bibitem{kaiser2019make}
F.~G. Kaiser, K.~Glatte, and M.~Lauckner, ``How to make nonhumanoid mobile
  robots more likable: Employing kinesic courtesy cues to promote
  appreciation,'' \emph{Applied ergonomics}, vol.~78, pp. 70--75, 2019.

\bibitem{thomas2018after}
J.~Thomas and R.~Vaughan, ``After you: doorway negotiation for human-robot and
  robot-robot interaction,'' in \emph{2018 IEEE/RSJ International Conference on
  Intelligent Robots and Systems (IROS)}.\hskip 1em plus 0.5em minus
  0.4em\relax IEEE, 2018, pp. 3387--3394.

\bibitem{Galati2022}
G.~Galati, S.~Primatesta, S.~Grammatico, S.~Macr{\`i}, and A.~Rizzo, ``Game
  theoretical trajectory planning enhances social acceptability of robots by
  humans,'' \emph{Scientific Reports}, vol.~12, no.~1, p. 21976, Dec 2022.

\bibitem{10341745}
C.~Cathcart, M.~Santos, S.~Park, and N.~E. Leonard, ``Proactive opinion-driven
  robot navigation around human movers,'' in \emph{2023 IEEE/RSJ International
  Conference on Intelligent Robots and Systems (IROS)}, 2023, pp. 4052--4058.

\bibitem{9022871}
B.~Gao and L.~Pavel, ``On passivity, reinforcement learning, and higher order
  learning in multiagent finite games,'' \emph{IEEE Transactions on Automatic
  Control}, vol.~66, no.~1, pp. 121--136, 2021.

\bibitem{samplingBasedAlgorithms_PRM}
S.~Karaman and E.~Frazzoli, ``Sampling-based algorithms for optimal motion
  planning,'' \emph{The International Journal of Robotics Research}, vol.~30,
  no.~7, pp. 846--894, 2011.

\bibitem{sucan2012the-open-motion-planning-library}
I.~A. {\c{S}}ucan, M.~Moll, and L.~E. Kavraki, ``The {O}pen {M}otion {P}lanning
  {L}ibrary,'' \emph{{IEEE} Robotics \& Automation Magazine}, vol.~19, no.~4,
  pp. 72--82, December 2012, \url{https://ompl.kavrakilab.org}.

\end{thebibliography}

\appendix
\textit{Probabilistic Roadmap:} 
We construct roadmaps -- each corresponding to a specific strategy in the navigation game -- using the PRM* algorithm from the OMPL library in the joint configuration space $\mathscr E^{N_r} \subset \mathbb R^{2N_r}$, where each configuration $\mathbf{x} = (p_1,\cdots,p_{N_r}) \in \mathscr E^{N_r}$ represents the 2D positions of all $N_r$ robots. A configuration $\mathbf{x}$ is considered valid if a minimum separation distance is maintained between every pair of robots:
$\|p_i - p_j\|_2 \geq 2r_{\text{robot}}, ~ \forall i \ne j$,
where $r_{\text{robot}}$ denotes the robot's radius.

Given the sets of origins and destinations of the $N_r$ robots, the PRM* planner finds an optimal path $(\mathbf x_1, \cdots, \mathbf x_N)$ by minimizing a cost function that combines the joint path length and a penalty for proximity between robots:
\begin{equation*}
\textstyle \sum_{i=1}^{N-1} \left\|\mathbf{x}_i - \mathbf{x}_{i+1}\right\|_2 + w_1 \sum_{i=1}^N \Phi_{\text{proximity}}(\mathbf{x}_i),
\end{equation*}
where $w_1 = 0.4$ is a weight parameter, and $\Phi_{\text{proximity}}$ is the proximity penalty term defined as:
\begin{equation*}
\textstyle \Phi_{\text{proximity}}(\mathbf{x}) = \sum_{i=1}^{N_r} \sum_{j=i+1}^{N_r} \frac{1}{\max\left( \|p_i - p_j\|_2 - 2r_{\text{robot}}, \, \varsigma \right)} . 
\end{equation*}
with a safety margin $\varsigma$ set to $0.1$ in our simulations. The path $(\mathbf x_1, \cdots, \mathbf x_N)$ is then linearly interpolated to generate a continuous trajectory $(p_i^{\footnotesize \text{PRM}}(t), v_i^{\footnotesize \text{PRM}}(t)), \, 0 \leq t \leq T_f$ for each robot~$i$, representing its position and velocity, while ensuring that the speed of every robot does not exceed $1~m/s$. 

\textit{Model Predictive Control:} 
We implement a third-order B-spline-based MPC to compute $(p_i (\tau), v_i (\tau)), \, t \leq \tau \leq t+T_{\footnotesize \text{MPC}}$ for each robot~$i$ at every time~$t$. Each robot~$i$ then moves according to the velocity $v_i(\tau), \, t \leq \tau \leq t+T_{\text{MPC}}$. The cost function and constraints are defined as follows:

\textbf{Cost function:}
\begin{align*}
    \textstyle \int_{\tau=t}^{t+T_{\footnotesize \text{MPC}}} ( \| p_i(\tau) - p_i^{\footnotesize \text{PRM}} (\tau)\|_2^2 + w_2 \| v_i(\tau) \|_2^2 ) \,\mathrm d\tau
\end{align*}

\textbf{Constraints:}
\begin{align*}
    &p_i(t) = p_{i,t}, \quad v_i(t) = v_{i,t} \\
  &\| p_i(\tau) - p_j^{\footnotesize \text{pred}}(\tau) \|_2 \geq 2 r_{{\footnotesize \text{robot}}}, \, j \neq i, ~ t \leq \tau \leq t+T_{\footnotesize \text{MPC}} \\
  &p_i(\tau) \in \mathscr E, ~ t \leq \tau \leq t+T_{\footnotesize \text{MPC}},
\end{align*}
where $p_i^{\footnotesize \text{PRM}}(\tau), ~ t \leq \tau \leq t + T_{\footnotesize \text{MPC}}$ represents the reference path of robot~$i$ determined by the precomputed PRM. The parameters $p_{i,t}$ and $v_{i,t}$ denote the position and velocity of robot~$i$ at time $t$, respectively, while $r_{\footnotesize \text{robot}}$ represents the robot's radius. Additionally, $p_j^{\footnotesize \text{pred}} (\tau), \, t \leq \tau \leq t + T_{\footnotesize \text{MPC}}$ denotes the predicted path of robot~$j$. If $j$ is one of robot~$i$'s game players, its predicted path is determined by the PRM. Otherwise, $p_j^{\footnotesize \text{pred}} (\tau)$ is set to robot~$j$'s current position for all $t \leq \tau \leq t + T_{\footnotesize \text{MPC}}$. The time horizon is set to $T_{\footnotesize \text{MPC}}=0.8$, and the weight in the cost function is chosen as $w_2 = 10^{-6}$.\footnote{The small weight $w_2$ is included solely to prevent abrupt changes in the resulting trajectory.}

\end{document}